\documentclass[final]{cvpr}

\usepackage{times}
\usepackage{epsfig}
\usepackage{graphicx}
\usepackage{amsmath}
\usepackage{amssymb}
\usepackage{multirow}

\usepackage{amsthm}
\theoremstyle{definition}
\newtheorem{definition}{Definition}
\theoremstyle{plain}
\newtheorem{theorem}{Theorem}
\theoremstyle{plain}
\newtheorem{lemma}{Lemma}
\theoremstyle{remark}

\makeatletter
\@namedef{ver@everyshi.sty}{}
\makeatother
\usepackage{tikz}
\usepackage[font=small,labelfont=bf]{caption}
\usepackage{booktabs}
\usetikzlibrary{positioning,arrows,calc}
\tikzset{
modal/.style={>=stealth',shorten >=1pt,shorten <=1pt,auto,node distance=1.5cm,
semithick},
world/.style={circle,draw,minimum size=0.5cm,fill=gray!15},
point/.style={circle,draw,inner sep=0.5mm,fill=black},
reflexive above/.style={->,loop,looseness=7,in=120,out=60},
reflexive below/.style={->,loop,looseness=7,in=240,out=300},
reflexive left/.style={->,loop,looseness=7,in=150,out=210},
reflexive right/.style={->,loop,looseness=7,in=30,out=330}
}

\usepackage{pgfplots}
\pgfplotsset{compat=1.16} 

\DeclareMathOperator*{\argmax}{argmax} 

\usepackage[pagebackref=true,breaklinks=true,colorlinks,bookmarks=false]{hyperref}

\def\cvprPaperID{****} 
\def\confYear{CVPR 2021}

\begin{document}

\title{Logically Consistent Loss for Visual Question Answering}
\author{Anh Cat Le Ngo \thanks{ Many thanks to Dr. Vuong Le and Mr. Thao Minh Le at A2I2, Deakin University for fruitful discussions and technical assistance}, Truyen Tran, Santu Rana, Sunil Gupta, Svetha Venkatesh}
\date{August 2020}

\maketitle


\begin{abstract}
Given an image, a back-ground knowledge, and a set of questions about an object, human learners answer the questions very consistently regardless of question forms and semantic tasks. The current advancement in neural-network based Visual Question Answering (VQA), despite their impressive performance, cannot ensure such consistency due to identically distribution (i.i.d.) assumption. We propose a new model-agnostic logic constraint to tackle this issue by formulating a logically consistent loss in the multi-task learning framework as well as a data organisation called family-batch and hybrid-batch. To demonstrate usefulness of this proposal, we train and evaluate MAC-net based VQA machines with and without the proposed logically consistent loss and the proposed data organization. The experiments confirm that the proposed loss formulae and introduction of hybrid-batch leads to more consistency as well as better performance. Though the proposed approach is tested with MAC-net, it can be utilised in any other QA methods whenever the logical consistency between answers exist.
\end{abstract}

\section{Introduction}


A Visual Question Answering task is correctly answering a question given visual clues. It comprises of understanding this visual signal structurally and the question semantically; then, systematically reasoning and inferring a probable answer. Inconsistency in answering related questions about the same object or subject given makes claims of truly intelligent Visual Question Answer (VQA) solutions \cite{hudson2019gqa}, questionable and brittle. State-of-the-art VQA methods \cite{hudson2018compositional} \cite{hudson2019gqa} \cite{le2020dynamic} \cite{tan2019lxmert} are often trained with batches of random questions regardless of its semantic relation to other questions. Therefore, such VQA solutions would struggle to understand underlying intents, occurring in other related questions. 
In addition, this may cause inconsistencies between answers if the questions are not parsed correctly or a reasoning system is unable to extract and chain question-relevant concepts and relations accurately. As an example, Figure 1 shows that they easily give wrong answer (\textcolor{red}{red} answer) to one of questions in three families of questions given corresponding images. One wrong answer puts the cloud of skeptism over whether a VQA can truly do visual reasoning task or it simply exhibits statistical correlation between questions and answers.

\begin{minipage}[t]{\columnwidth}
    \centering
    \begin{tabular}[\columnwidth]{{lp{4.3cm}}}
        \multirow{4}{*}{\includegraphics[width=3cm]{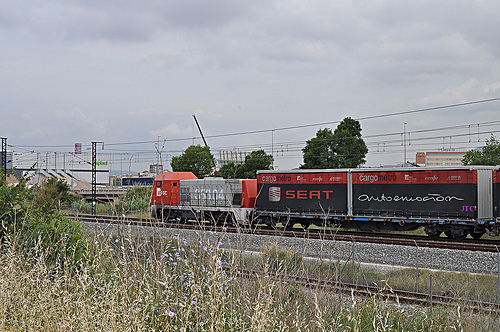}} 
        & \footnotesize{1. Does the sky look cloudy? \textcolor{green}{yes}/\textcolor{red}{no}} \\ 
        & \footnotesize{2. Is the gray sky cloudy or clear? \newline \textcolor{green}{cloudy}/\textcolor{red}{clear}}  \\ 
        & \footnotesize{3. Is the sky both sunny and gray? \textcolor{green}{no}/\textcolor{red}{yes}} \\ 
        & \footnotesize{4. Does the gray sky look clear? \textcolor{green}{no}/\textcolor{red}{yes}} \\[.25cm]
    
        \multirow{5}{*}{\includegraphics[width=3cm,height=2.7cm]{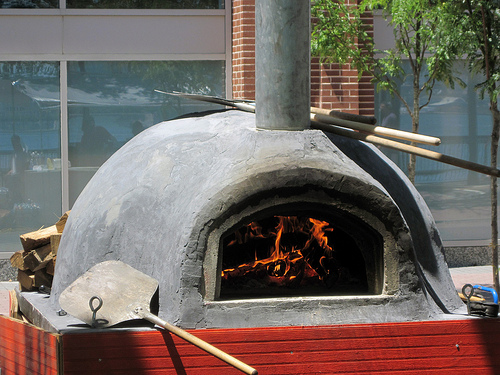}}
        & \footnotesize{1. What color is the pipe that \newline is above the oven? \textcolor{green}{silver}/\textcolor{red}{black}} \\
        & \footnotesize{2. Does that pipe look silver? \textcolor{green}{yes}/\textcolor{red}{no}} \\
        & \footnotesize{3. Is the pipe black?} \textcolor{green}{no}/\textcolor{red}{yes} \\
        & \footnotesize{4. Which color is the pipe made of metal, silver or black? \textcolor{green}{silver}/\textcolor{red}{black}} \\
        & \footnotesize{5. Is the metal pipe wide and red? \textcolor{green}{no}/\textcolor{red}{yes}} \\[.25cm]
        
        \multirow{6}{*}{\includegraphics[width=3cm,height=3.7cm]{figures/2410128.jpg}} 
        & \footnotesize{1. What is the size of the banana to the right of the other banana? \textcolor{green}{small}/\textcolor{red}{big}} \\
        & \footnotesize{2. Is the banana on the tree small and ripe? \textcolor{green}{no}/\textcolor{red}{yes}} \\
        & \footnotesize{3. Is the banana on the tree small or large? \textcolor{green}{small}/\textcolor{red}{large}} \\
        & \footnotesize{4. Does the green banana look small? \textcolor{green}{yes}/\textcolor{red}{no}} \\    
        & \footnotesize{5. Does the banana that looks unripe have large size? \textcolor{green}{no}/\textcolor{red}{yes}} \\
        & \footnotesize{6. How big is the unripe banana? \textcolor{green}{small}/\textcolor{red}{big}} \\
    \end{tabular}
    \captionof{figure}{Three examples from GQA dataset: (top image) image-id: 2363334, (middle image) image-id: 2391016 and (bottom image) image-id: 2410128 with green correct answers / red wrong answers. All questions are about the same objects given in correspondent images and all answers have to be in green to be logically consistent. Otherwise, one or more answers will be contradictory to the others in the same family.}
    \label{fig:intro}
\end{minipage}

Two main causes of inconsistent VQA solutions are the inability to comprehend rephrased queries with different linguistic forms and the ignorance of semantically related queries. Existing works to enhance consistency form two lines of research: introducing more consistent samples \cite{shah2019cycle}\cite{ray2019sunny}\cite{selvaraju2020squinting}\cite{do2020multiple}\cite{gokhale2020mutant} and using logic in parsing textural questions into logical queries \cite{gokhale2020vqa} \cite{amizadeh2020neuro}.  These works focus on a pair of consistent samples or logically parse an individual question; however, they ignore semantic structures between related questions about the same subject / object in an image. Different from above solutions, our work enforces consistency across families of questions with unique semantic tasks. In particular, we make use of the recently introduced GQA data set, where each question is assigned a semantic task, and the relationships between questions are specified. These extra pieces of knowledge are utilized to form a logic-induced loss at training time in a multi-task learning setting. At test time, however, the trained model does not assume the availability of these labels. For an arbitrary question posed by the end user, the model recognizes the semantic task and provides an answer instead. This strategy of "train jointly - test independently" is practical because questions at test time can be random in order and in time.

More specifically, the semantic relations between these tasks are pre-defined by entailment graphs (c.f. Fig. 14 in Supplementary Material of \cite{hudson2019gqa}). In line with Tarski et al. \cite{tarski1994introduction} who define consistency in the context of deductive logic that evidences should not generate entailed contradiction, we introduce a logic rule for consistency: for all questions in a family if semantic tasks of two related questions are correctly predicted, their respective answers are both correctly predicted as well. We introduce our method, that is, applying T-norm fuzzy logic to translate the stated logic into a consistent-augmented loss function which can be used for a wide-rage of VQA solutions. In this work, we are experimenting this novel loss function with MAC-net \cite{hudson2018compositional} as our chosen VQA machine but we wish to emphasize that the solution is general and agnostic of the underlying backbone model.

\section{Related works}

The literature of improving consistency can be organized into three groups of strategy: data augmentation, multitask learning and logically reasoning. These three strategies are often utilized together in following related works for consistency and generalization improvement of VQA solutions.

\textbf{Data Augmentation} This approach hypothesizes that inconsistent VQA solution is due to shortage of related samples in which questions are rephrases of other questions, closely linked by semantic tasks or loosely connected by their meaning. This paragraph documents few efforts of creating or generating such data set but does not provide a comprehensive survey of this research approach. Shah et al \cite{shah2019cycle} collect a large-scale data set of rephrasing questions for images of VQA 2.0 data set and propose a cycle-consistent training scheme to minimize loss / inconsistency between a pair of question, answer and its generated counterparts. Ray et al. \cite{ray2019sunny} and Selvaraju et al. \cite{selvaraju2020squinting} separately collect 'Commonsense-based Consistent QA' and 'VQA-introspect' of which semantically related questions and answers are annotated by humans. While Ray et al. \cite{ray2019sunny} use data-augmentation approach to enforce consistency between pairs of questions and answers, Selvaraju et al. \cite{selvaraju2020squinting} introduce attention loss, inconsistency or difference between grounded visual areas, for reasoning about answers between a main question and a sub question. Though we do not create or generate additional data in this work, we propose organization of samples into families in which questions are about the same subject or object in images. This reorganization of data gives strong training signals for consistency of VQA solutions.

\textbf{Multitasks Learning} Multi-task learning improves generalization of machine learning solutions by leveraging the domain-specific information contained in training signals of related tasks, eloquently put by Caruana \cite{caruana1998multitask}. More practically, Ruder \cite{ruder2017overview} defines multi-task learning as optimizing more than one loss functions. Extra tasks beside correct classification of answers are implemented for improving consistency and generalization of VQA solutions. For example, Do et al. \cite{do2020multiple} aim to maximize statistical correlation between question-types and answers by penalizing incoherence between question-types and answers. Meanwhile, Gokhale et al. \cite{gokhale2020mutant} formulate an extra training objective called Noise Contrastive Estimation over cross-modal features and answer embedding on a shared manifold. Saha \cite{saha2018complex} combines a task of answering simple factual questions and learning a knowledge graph w.r.t. to datasets of conversational questions \& answers. These conversational questions may loosely be defined as families of questions but they are not grounded to any visual information. Shevchenko et al. \cite{shevchenko2020visual} add reconstruction of embedded answer vectors as additional training objective and incorporate distance between reconstructed vectors and pre-trained word-embedding dictionary into their inference stage. Despite popularity of multi-task approaches in VQA solutions, there is lack of utilising semantic tasks and more importantly known semantic relations between these tasks / intents of questions. 

\textbf{Logical Reasoning} Consistent samples contain questions and answers, fitting into prior knowledge or knowledge graph of logical relations between semantic tasks. Therefore, improving logical reasoning is strongly considered a potential solution for improving consistency. Hudson et al. \cite{hudson2019learning} propose Neural-State Machine, carrying out soft-attention reasoning over scenegraphs with functional programs, parsed from textual questions. Instead of parsing visual information as scene-graphs, Amizadeh et al. \cite{amizadeh2020neuro} interpret the information as First-Order Logic statement and develop differentiable First-Order-Logic framework to learn optimal attention module for inferring correct answers w.r.t. logic queries, parsed from questions. Gokhale et al .\cite{gokhale2020vqa} conduct detailed analysis of several VQA's models on their generated, logically composed questions and proposes dedicated attention module for parsing logical connectives from questions. These mentioned works explicitly focus on logic existing in either visual or lingual modality but none explores logical links between related questions, often existed in real-life VQA scenarios.


\section{Method}

In the proposed approach, we assume there existing semantics tasks and logical relations between questions i.e.  entailment graphs as seen in the recent GQA data set \cite{hudson2019gqa}. In this data set, there are $\mathcal{T}=48$ semantic tasks and their relational semantic logic formulae, listed in Table \ref{tab:semtsk} and Table \ref{tab:semlog} in the Supplementary Materials respectively. Table \ref{tab:family-batch-1} shows an example of semantic tasks and their relations along side correspondent questions and answers. This work aims to utilise this prior knowledge to find consistent answers by correctly recognising semantic tasks and enforcing their logical entailment. In other words, we enforce a consistency by maximising satisfaction of known structural entailment between semantic tasks and answers. This is equivalent to minimizing our proposed consistency-augmented loss derived from such logic. In Subsection \ref{subsec:prelim}, we formally define answer and semantic task recognition problems. Then, we describe structures in families of questions and mention availability of such structures in GQA's data set in Subsection \ref{subsec:fambat}. Subsection \ref{subsec:conlog} contains formulation of our proposed consistent logic and how T-norm is used to translate such logic into a continuous loss function.
\subsection{Prelims}
\label{subsec:prelim}
Visual Question Answer task is finding the most probable answer $a_i$ given embedded vectors $\mathbf{q}$ of textual questions and visual images $\mathbf{K}$ over a set of possible answers $\mathcal{A}$, defined in Equation \ref{subsec:prelim:eqn1}. 
\begin{align}
    a_i &= \argmax_{a_i \in \mathcal{A}} \mathbf{p^a} = \argmax_{a_i \in \mathcal{A}} P(Answer = a_i \vert \mathbf{q}, \mathbf{K}) \label{subsec:prelim:eqn1}
\end{align}
Whenever possible we also propose an inclusion of an auxiliary task called Question Semantic-Task Recognition which is finding the most probable semantic-task $Task$ labels of given questions $\mathbf{q}$ over a set of 48 possible semantic labels $\mathcal{T}$. It is formally defined in Equation \ref{subsec:prelim:eqn2}.
\begin{align}
    t_m &= \argmax_{t_m \in \mathcal{T}} \mathbf{p^t} = \argmax_{t_m \in \mathcal{T}} P(Task = t_m \vert \mathbf{q}) \label{subsec:prelim:eqn2}    
\end{align}
We assume that textual questions and visual images are transformed to vector representation. In our actual implementation, we employ bi-LSTM and CNN neural network respectively, as suggested in \cite{hudson2018compositional}. Henceforth, we call this transformation as input transformation function and denote it as $f_{ip}$, defined as follows.
\begin{align}
    \mathbf{q}, \mathbf{K} &= f_{ip}(Question, Image) \label{subsec:prelim:eqn0}
\end{align}
The function $f_{ans}$, mapping the embedded vectors $\mathbf{q}, \mathbf{K}$ to answer probabilities $\mathbf{p^a} \in \mathcal{R^{\mathcal{\vert A \vert}}}$, is defined in Eqn. \ref{subsec:prelim:eqn:fans} and its parameters may be trained and approximated by several types of neural networks e.g. MAC \cite{hudson2019gqa}, LSTM etc. 
\begin{align}
    \mathbf{p^a} &= f_{ans}(\textbf{q}, \textbf{K}) \label{subsec:prelim:eqn:fans} \\ 
    \mathbf{p^t} &= f_{tsk}(\textbf{q}) \label{subsec:prelim:eqn:ftsk}
\end{align}
Similarly, a simple neural network architecture may approximate the function $f_{tsk}$, Eqn. \ref{subsec:prelim:eqn:ftsk} for probabilities across possible semantic tasks $\mathbf{p^t} \in \mathcal{R}^{\mathcal{\vert T \vert}} $ given a question $\mathbf{q}$.

Finally, relations between semantic tasks, as illustrated in Fig \ref{fig:family-batch-1}, are written as First-Order Logic (FOL) functions. The full list of relations are listed in Table \ref{tab:semlog} of the Supplementary Material. With respect to these relations, several potential logic formulae $\psi$ for consistent answers can be hypothesized and following are two examples, 
\begin{align}
    \exists x_1 \exists x_2 & \big(\text{verifyGlobalTrue}(x_1) \Rightarrow \text{verifyGlobalFalse}(x_2)\big) \nonumber \\
    \Rightarrow & \big(a_1 \Leftrightarrow \text{yes} \otimes a_1 \Leftrightarrow \text{no}\big) \nonumber \\ 
    \exists x_1 \exists x_2 & \big(a_1 \Leftrightarrow \text{yes} \otimes \text{verifyGlobalTrue}(x_1)\big) \nonumber \\
    \Rightarrow & \big(a_2 \Leftrightarrow \text{no} \otimes t_1=\text{verifyGlobalTrue}(x_2)\big) \nonumber
\end{align}
where $x_1, x_2$ are two sample data; $a_1, a_2$ are predicted answers for these samples. VerifyGlobalTrue($x_1$) and verifyGlobalFalse($x_2$) represent truth degrees or probabilities of truth about this statement: semantic tasks of $x_1$ and $x_2$ are "verifyGlobalTrue" and "verifyGlobalFalse" respectively. The logic connectives "exists" $\exists$, "conjunction" $\otimes$, "residual-imply" $\Rightarrow$ and bi-residuum $\Leftrightarrow$ are defined by T-norm and Fuzzy Logic theory. This theory provides a systematic method of translating FOL formulae $\psi$ of consistency to constraint satisfaction functions $f_{\psi}$ and subsequently a consistency-augmented loss function $\mathcal{L}_{\psi}$. We can estimate parameters of neural networks with training s.t. this loss function is minimized. In other words, predictions, produced by these neural networks, optimally satisfy the consistent logic constraints. Interestingly, supervised tasks i.e. classification of answers and semantic task can also be written in FOL formulae as $\forall x \quad p^x \Leftrightarrow \text{label}$. This observation helps streamline integration of any logic loss in the existed supervised loss function. Further details about T-norm generators, Fuzzy Logic can be found in the Appendix \ref{apd:tfl} and \cite{giannini2019learning}.

\subsection{Semantic Structures in Families of Questions}
\label{subsec:fambat}

\begin{figure}[htbp]
    \centering
    \begin{tikzpicture}[modal]
        \node[world,label={above:verifyGlobalTrue}] (verifyGlobalTrue) {};
        \node[world, label={above:verifyGlobalFalse}] (verifyGlobalFalse) [below left=of verifyGlobalTrue] {};
        \node[world,label={below:queryGlobal}] (queryGlobal) [below right=of verifyGlobalTrue] {};
        \node[world, yshift=.5cm, label={right:chooseGlobal}] (chooseGlobal) [below=of verifyGlobalFalse] {};
        \path[->] (verifyGlobalTrue) edge[] (verifyGlobalFalse);
        \path[<->] (verifyGlobalTrue) edge[] (queryGlobal);
        \path[->] (verifyGlobalFalse) edge[] (chooseGlobal);
    \end{tikzpicture}
    \caption{Entailment graph about global attributes of an image}
    \label{fig:family-batch-1}
\end{figure}
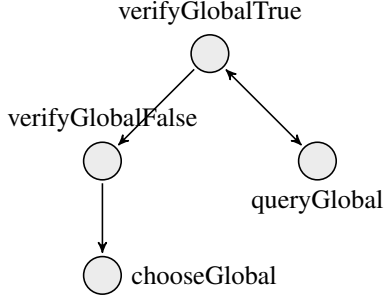

\begin{table}[htbp]
    \centering
    \begin{tabular}[b]{|p{1.9cm}|p{3cm}|l|}
        \hline
        Semantic \newline Task & Questions & Answers \\
        \hline
        Verify \newline GlobalTrue  &  Is this an image  of a beach? & yes \\ \hline
        Verify \newline GlobalFalse  &  Is it a zoo ? & no \\ \hline
        Query \newline Global & Which place is  it? & beach \\ \hline
        Choose \newline Global & Is it a beach or  a farm? & beach \\ \hline
    \end{tabular}
    \caption{Examples of questions and answers for Figure \ref{fig:family-batch-1} }
    \label{tab:family-batch-1}        
\end{table}

Questions and answers about an image are usually related as they are grouped together as families of questions about the same subject or object in that image. In other words, significant relational structure exists in training VQA machines, which are often ignored due to popularity of minibatches with i.i.d. samples. Assumed relational structures are fixed and known, Hidden Markov Logic \cite{kok2009learning} and Problog \cite{de2007problog} are two exemplary frameworks for incorporating this structured prior knowledge into probabilistic learning and inference solutions. When these structures are not available, structure learning needs carrying out over a large space of hypotheses by inductive logic programming or conditional random field. In this work, we hypothesize integration of semantically relational structure between a family of questions would promote consistency and propose hybrid batches, including both families of questions and randomly selected ones, in place of i.i.d. minibatches for training VQA machines.

To create a family of questions, we leverage semantic strutures in GQA dataset \cite{hudson2019gqa} as each question this dataset has a field 'argument'. Hence, it enables usage of SQL for grouping related questions about the same subject / object existing in the visual knowledge - an image. In addition, these questions have specific sub-fields 'semantic' and 'entailed' which label semantic-tasks and describe their relationship to other questions in the same family. Furthermore, fixed relational structures of question families are provided in GQA dataset as entailment graphs between semantic tasks (c.f. Fig 14 in Supplementary material of \cite{hudson2019gqa}). Figure \ref{fig:family-batch-1} illustrates a structure of a question family in GQA dataset. Table \ref{tab:family-batch-1} shows a family batch of six questions with unique semantic tasks, logically related to each other w.r.t. entailment graphs in Fig. \ref{fig:family-batch-1}.  A family may have questions corresponding to semantic tasks and their relations e.g. solid lines if Fig. \ref{fig:family-batch-1} but usually missing links appear e.g. dashed links in Fig. \ref{fig:family-batch-2} in the Supplementary Material.

\subsection{Consistent Logic}
\label{subsec:conlog} 
Given a data set $\mathcal{X}$, training a machine learning solution is a process of finding a function whose output w.r.t. input $\mathcal{X}$ satisfying a set of constraints, representing objectives and knowledge about the problem to be solved. A single-task supervised learning is a single point-wise constraint problem $p$ with a loss function $\mathcal{L}(\mathcal{X},p)$. Multi-task learning is a multiple point-wise constraint problem $\mathbf{p}=\{p_1,\ldots,p_J\}$ with $J$ different tasks \cite{ruder2017overview}. A logic-constrained multitask learning is a multi-task learning with a knowledge base $\mathcal{KB}=\{\phi_1,\ldots,\phi_H\}$ of $H$ different logical relations between different $J$ task functions and these logical relations $\phi$ are First-Order Logic (FOL) formulae. Given the task definition, learning is finding suitable parameters for function approximators - neural networks - to satisfy all supervised tasks $J$ as well as their logical relations stated in $\mathcal{KB}$. Overall, the loss function for a logic-based multi-task learning is written as follows.
\begin{equation}
    \mathcal{L}(\mathcal{X},\mathcal{KB},\mathbf{p})=\sum_{i=0}^{J+H} \beta_i L\Big(f_{i}\big(\mathcal{X},\mathcal{KB},\mathbf{p}\big)\Big)
\end{equation}
Here are two considered supervised tasks $f_{ans}$ and $f_{tsk}$ which are classifications of answers $ans$ and semantic-task $tsk$ of a question $\textbf{q}$ about an image $\textbf{K}$. 
An input to both functions comes from an input unit $f_{ip}$ described in Sub-section \ref{subsec:prelim};therefore, an output of mapping function for probability distribution of answers is written as $(f_{ans} \circ f_{ip})(x)$ and similarly an output of $(f_{tsk} \circ f_{ip})(x)$ is probability distribution over semantic tasks. 

There are 1845 distinctive answers and 48 different semantic-task labels for questions in the GQA dataset; hence, outputs of $(f_{ans} \circ f_{ip})(x)$ and $(f_{tsk} \circ f_{ip})(x)$ are logit vectors of $N_a$ and $N_t$ probabilities $[p^{a_1},p^{t_2},\ldots,p^{t_{1845}}] \in \mathbb{R}^{1845}$ and $[p^{t_1},p^{t_2},\ldots,p^{t_{48}}] \in \mathbb{R}^{48}$, normalised by respective softmax layers. These probabilities $p^{a_i}, p^{t_j}$ represent truth degree for which a question $x$  has an answer $a_i$ and a semantic-task $t_j$. 
\begin{align}
\big[p^{a_1}(x),p^{a_2}(x),\ldots,p^{a_{1845}}(x)\big] &= (f_{ans} \circ f_{ip})(x) \\
\big[p^{t_1}(x),p^{t_2}(x),\ldots,p^{t_{48}}(x)\big] &= (f_{tks} \circ f_{ip})(x)
\end{align}
For supervised learning tasks i.e. classifications of answers and semantic tasks, labels $A^{i}(x),T^{m}(x)$ are known for training samples; the knowledge-base rules of such supervision constraints are written as follows.
\begin{align}
    \psi_{ans}(x) &:= \forall x \varphi_{ans}(x) := \forall x A^{i}(x) \Leftrightarrow p^{a_i}(x) \\
    \psi_{tsk}(x) &:= \forall x \varphi_{tsk}(x) := \forall x T^{m}(x) \Leftrightarrow p^{t_m}(x)
\end{align}

Besides supervised-learning tasks, this paper proposes the logical rules $\mathcal{KB}$ as consistency constraints among questions of a family-batch. Labels of two different questions $x_1,x_2$ are ${\big(A^i(x_1),T^m(x_1)\big),\big(A^j(x_2),T^n(x_2)\big)}$, shorten to ${\big(A^i_1,T^m_1\big),\big(A^j_2,T^n_2\big)}$ henceforth. Similarly predictive probabilistic output of answers and tasks for $x_1, x_2$ are also shorten to $(p^{a_i}_1,p^{t_m}_1)$ and $(p^{a_j}_2,p^{t_n}_2)$. The first potential FOL logic formula for such consistency is 
\begin{align}
\varphi(x_1,x_2) = & \Big[ \big( T^m_1 \Leftrightarrow p^{t_m}_1 \big) \otimes \big( T^n_2 \Leftrightarrow p^{t_n}_2 \big)\Big] \nonumber \\ 
\Rightarrow & \Big[ \big( A^i_1 \Leftrightarrow p^{a_i}_1 \big) \otimes \big( A^j_2 \Leftrightarrow p^{a_j}_2 \big) \Big]
    \label{lgceqn:c_rule}
\end{align}
As the logic involves two inputs and the constraint should be enforced against all pairs of $x_1,x_2$, the appropriate quantifier for is $\forall x_1 \forall x_2$. The above logic formula states that correct prediction of semantic tasks of $f_{tsk}$  implies a higher chance of getting correct answers with $f_{ans}$ for related questions. Despite outputs $p^{a_i}, p^{t_m}$ of different function $f_{ans}, f_{tsk}$, they are results of inputs from the same input unit $f_{ip}(x)$ for a given sample $x$, described in Sub-section \ref{subsec:prelim} 
. Hence, this logic loss says that learnt representations of a question pair for accurately identifying semantic tasks implies good representation for finding out answers of the question pair. As the questions are organised in a family-style in this work, the $f_{ans}$ and $f_{tsk}$ are trained to provide consistent answers to a set of question about the same subject / object. 

In addition, there are further entailment relations between semantic tasks of questions in the same family, described in Sub-section \ref{subsec:fambat}. Such entailment relations can be translated into logic-loss as well. In GQA data set, there are 45 different possible entailed relations between a pair of different questions $x_1, x_2$, listed in a supplementary Table \ref{tab:semlog}. For each entailment, the logical relation between a question pair is either residual-imply $\Rightarrow$ or bi-residuum $\Leftrightarrow$. Though there are 45 different logical rules for each pair of question i.e. $ R^k:= p^{t_m}_1 \square p^{t_n}_2$ where $\square = \{\Rightarrow,\Leftrightarrow\}$ and $k \in [1,45]$, there is only one direct relation for a pair of two different questions w.r.t. the entailment graph. Therefore, logical relations between 45 $R_k$ are $or - \vee$ operators. These entailment logic formulae may replace the conjunction of semantic-task prediction of Eqn \ref{lgceqn:c_rule} as follows.
\begin{align}
 \varphi(x_1,x_2) &:= \Big[ \big( R^1 \vee \ldots \vee R^{45} \big) \Big] \\
 &\Rightarrow \Big[ \big( A^i_1 \Leftrightarrow p^{a_i}_1 \big) \otimes \big( A^j_2 \Leftrightarrow p^{a_j}_2 \big) \Big] \nonumber
    \label{lgceqn:e_rule}
\end{align}
\subsection{Consistent Loss}
\label{subsec:conlos}
Giannini et al. \cite{giannini2019learning} suggests T-Norm theory,  continuous and non-increasing generator $g$ is utilised to systematically translate above FOL logics into loss functions. The first step is converting quantifiers $\forall x$ and $\forall x1,x2$ of $\varphi(x)$ and $\varphi(x_1,x_2)$. As the quantifiers means applying a constraint $\varphi$ for all samples of $x$ or pairs $x_1,x_2$, its role have to be interpreted as follows,
\begin{align}
    \psi_1(x)&=  \forall x \varphi(x) \simeq \varphi_1(x^1) \otimes \ldots \otimes \varphi(x^N)  \\
    \psi_2(x_1,x_2)&= \forall x_1 \forall x_2 \varphi_2(x_1,x_2) \\ 
    &\simeq \varphi(x^1_1,x^1_2) \otimes \ldots \otimes \varphi(x^N_1,x^N_2) \nonumber
\end{align}
where $N$ is the number of samples in $\mathcal{X}$. As T-norm is defined in Definition \ref{def:tnorm-1} in the Supplementary Material with an additive generator $g$, the generated continuous truth degree of $\psi(x) = \forall x \varphi(x) $ and $\psi(x_1,x_2) = \forall x_1 \forall x_2 \varphi(x_1,x_2)$ are given as follows.
\begin{align}
    f_{\psi_1}(\mathcal{X},\mathbf{p}) &= g^{-1}\bigg( \min \bigg\{ g(0^{+}), \sum_{x\in\mathcal{X}} g(f_{\varphi}(x,p)) \bigg\} \bigg) \nonumber \\
    f_{\psi_2}(\mathcal{X},\mathbf{p}) &= g^{-1}\bigg( \min \bigg\{ g(0^{+}), \sum_{x_1\in\mathcal{X}} \sum_{x_2\in\mathcal{X}} g(f_{\varphi_2}(\mathbf{x},\mathbf{p})) \bigg\} \bigg) \nonumber
\end{align}
Then, the above truth degree function is mapped to a loss function by applying a decreasing function upon its output. A generator $g$ is a decreasing function with lower bound $g(1)=0$, which a natural choice for mapping a truth function to a loss function. Hence, the loss function of $\psi_1$ and $\psi_2$ are
\begin{align}
    L_{\psi_1} &= 
    \begin{cases}
        \min \bigg\{ g(0^{+}), \sum\limits_{x\in\mathcal{X}} g(f_{\varphi}(x,p)) \bigg\} &\text{if } g(0^+) < +\infty \\
        \sum\limits_{x\in\mathcal{X}} g(f_{\varphi}(x,p)) &\text{if } g(0^+) = +\infty
    \end{cases} \nonumber
    \\
    L_{\psi_2} &= 
    \begin{cases}
        \min \bigg\{ g(0^{+}), \sum\limits_{x_1,x_2\in\mathcal{X}}  g(f_{\varphi}(\mathbf{x},\mathbf{p})) \bigg\} &\text{if } g(0^+) < +\infty \\
        \sum\limits_{x_1\in\mathcal{X}} \sum\limits_{x_2\in\mathcal{X}} g(f_{\varphi}(\mathbf{x},\mathbf{p})) &\text{if } g(0^+) = +\infty
    \end{cases}  \nonumber  
\end{align}
Regarding to point-wise supervised losses of classifying answers and semantic-tasks, the generated loss function of a sample $x$ are
\begin{align}
    \varphi A^i \Leftrightarrow p_{a_i} \Rightarrow L_{ans}(\mathcal{X},p_{a_i}) &= \sum_{x\in\mathcal{X}} \vert g(A^i) - g(p^{a_i}) \vert \nonumber \\
    \varphi T^m \Leftrightarrow p_{t_m} \Rightarrow L_{tsk}(\mathcal{X},p_{t_m}) &= \sum_{x\in\mathcal{X}} \vert g(T^m) - g(p^{t_m}) \vert \nonumber
\end{align}
Only positive answers and tasks are provided in QQA training phases; hence $g(A^i) = 0$ and $g(T^m) = 0$; therefore, the supervised losses become
\begin{align} 
L^{+}_{ans} = \sum_{x\in\mathcal{X}} g(p^{a_i}), L^{+}_{tsk} = \sum_{x\in\mathcal{X}} g(p^{t_m}) \nonumber
\end{align}
If functions $g(x) = 1 -x$ or $g(x) = -\log(x)$ is chosen as generators, it corresponds to Lukasiewicz or Product T-norm and we have respective loss following supervised functions, respectively:
\begin{align} 
L^{+}_{ans} = \sum_{x\in\mathcal{X}} (1-p^{a_i}) &, L^{+}_{tsk} = \sum_{x\in\mathcal{X}} (1-p^{t_m}) \nonumber \\
L^{+}_{ans} = \sum_{x\in\mathcal{X}} -\log(p^{a_i}) &, L^{+}_{tsk} = \sum_{x\in\mathcal{X}} -\log(p^{t_m}) \nonumber
\end{align}
Interestingly, Lukasiewicz T-norm of supervised-learning logic translates to $L_1$ loss; meanwhile, cross-entropy loss is generated by Product T-norm of supervised-learning logic. The deliberate choice of using the T-norm generator $g$ for mapping truth-degree formula $f_{\varphi}$ to its loss function eliminates an occurrence of $g^{-1}$ in the final loss function is called the \textit{simplification property}. Giannini et al. \cite{giannini2019learning} generalizes and proves this property for a restricted list of logical connectives in the Lemma \ref{lem:sim}.
\begin{lemma}
Any formula $\varphi$, whose connectives are restricted to {$\wedge, \vee, \otimes, \Rightarrow, \sim, \Leftrightarrow$}, has the simplification property.
\label{lem:sim}
\end{lemma}
The simplification property allows elimination and computation of all pseudo-inverse $g^{-1}$ in any grounded formula with connectives from the limited list. Furthermore, it allows general implementation of n-ary ($n \geq 2$) t-norms i.e. universal quantifiers $\forall, \exists$. Equipped with this lemma, the consistency logic formula \ref{lgceqn:e_rule} is mapped to its loss function below for a strict T-norm generator $g$.
\begin{align}
    L_{\psi} = \sum_{x_1,x_2 \in \mathcal{X} } \max \bigg\{ 0, g(p^{a_i}) + g(p^{a_j}) \nonumber - g\big( \max_{k\in[1,45]} \big\{p^{R_k}\big\} \big) 
    \bigg\}
\end{align}
In case of $g(x) = -\log x$ - generator of product T-norm, the loss function becomes
\begin{align}
    L_{\psi} = \sum_{x_1 \in \mathcal{X}} \sum_{x_2 \in \mathcal{X}} \max \bigg\{ 0, &-\log\big(p^{a_i}\big) -\log\big(p^{a_j}\big)  \\ + \log\big( \max_{k\in[1,45]} \big\{p^{R_k}\big\} \big) 
    &+ \big(\log(p^{t_m}) + \log(p^{t_n})\big) \bigg\} \nonumber
\end{align}

Overall, the loss function used for the logically consistent visual question answer is written as follows.
\begin{equation}
    \mathcal{L} = \beta L_{\psi} + L^{+}_{ans} + L^{-}_{tsk}
\end{equation}
where $\beta$ is a weight for the constrain-augmented loss while weights for both supervised tasks are set to 1.0. Higher value of $\beta$ requires higher degree of satisfaction for consistency constraint. This weight is set to a constant value in this work; however, it can be learnt from data as suggested in \cite{marra2019integrating}. The multi-task training or computation of multiple losses are only required in the training phases of this proposal. Therefore, families of questions or batches of related questions are only necessary during training stages. In validation and inference stages, the proposed consistent VQA solution can produce answers for both random and related questions.

The consistency-augmented loss can be alternatively interpreted as an explicit inductive bias since this consistent-induced logic formulae is a set of rule for VQA machines to infer outputs of unseen inputs. Ruder \cite{ruder2017overview} observes that inductive bias in multi-task learning is commonly $l_1$ regularization. The proposed loss $L_{\psi}$ is formulated $l_1$ loss with a specific generator class $g$ ( i.e. The  Schweize-Sklar family ) and its corresponding parameter ( $\lambda = 1$ ). The $l_1$ regularization leads to a preference for sparse solutions in the concerned learning tasks, heavily favoring the consistent set of answers and semantic tasks than the rest. Cross-entropy losses are employed for learning the other tasks $L^{+}_{ans}$ and $L^{+}_{ans}$ by selecting an appropriate generator ( i.e. from the Schweize-Sklar family ) with its parameter value ( $\lambda = 0$).

\section{Experiments}
\label{sec:exp}
\subsection{Data \& Arrangement for Training}
This paper aims to train a VQA solution which can logically consistent answer when input questions are about the same subject / object w.r.t. to the same image. Though randomly arranged batch of samples are predominantly used in normal settings, they are not suitable for the above training purpose. Therefore, we use SQL to regroup questions into families of questions w.r.t. their 'entailed' and 'argument' ( fields available in the data structure of a GQA sample ) relationship instead of batches of random samples. Each family is generated such that each member question has a unique task for avoiding two different questions in wordings but with the same semantic task. As the result, we have families with number of samples in a range from 1 to 6 questions with unique semantic tasks. These family of questions are later used for training $f_{ans}$ and $f_{tsk}$ by minimising a combined supervised losses $L^{+}_{ans}$ and $L^{+}_{tsk}$ as well as the proposed consistent-augmented loss $L_{\psi}$.

The size of each family, 1-6 questions, are too small for forming a reasonable training batch. In this paper, we propose a hybrid batch of which the first few samples are from a family and the rest are randomly selected samples from 'balanced' subset of GQA dataset. This hybrid approach allows a feasible comparison between a normal training approach with purely random batches, and more structured hybrid data employed in this work. Remarkably, the hybrid batch reflects more realistic scenarios of how human learn to answer a group of related questions given visual clues. Hence, training with hybrid batches probably benefits performances of VQA solutions as they can learn these underlying logical relations and produce appropriate non-conflicting answers.

\subsection{Implementation Details}
In this work, the Horovod framework \cite{sergeev2018horovod} is utilized for distributively training VQA solutions in parallel to speed up training time and deal with a large number of training data. The whole training data are equally spited into 16 partitions with randomly repeated samples as the total number of samples are not divisible by 16. Then, the training process is started with 16 workers, utilising either 16-GPU DGX-2 server or two 8-gpu DGX-1 servers. On each GPU, a Tensorflow implementation of $f_{ip}$, $f_{ans}$ and $f_{tsk}$ is initialized with identical initial parameters across workers on all GPUs but with one of 16 data partitions. On each GPU, a batch-size of 16 samples include a family of $n \in [1,6]$ samples and $16-n$ randomly selected samples from 'balanced' subset of GQA data set. Effectively, the training batch-size for the whole parallel process is 256 samples as the gradient vectors for updating model parameters are synchronized by all-reduce algorithm across all 16 workers despite each work carries out forward and backward computation on separate portion of training data. Adasum all-reduce algorithm is utilised here; further details about this algorithm and Horovod in general can be found in \cite{sergeev2018horovod}. Hudson et al. used a batch-size of 64 and learning rate 1e-4 in \cite{hudson2018compositional} and a default batch-size of 128 and learning rate 3e-4 in a released code for \cite{hudson2019gqa}. As we use an effective batch-size of 256 in this work and Adasum all-reduced algorithm, the effective learning rate is chosen to be 1e-4. Our model is trained by 60 epochs without early stopping and the rest of training configurations suggested in \cite{hudson2018compositional}, i.e. word vector dimensions 300, variational dropout. Importantly, a difference in NN architecture between MAC architecture used in \cite{hudson2019gqa} and following experiments is usage of soft-max layers instead sigmoid layers at the final step of $f_{ans}$ and $f_{tsk}$. Outputs of soft-max layers have values in range $[0,1]$ and their sum is 1.0, which are more suitable for probabilistic interpretation of truth-degree, necessary for logic-loss computation in subsequent steps. 

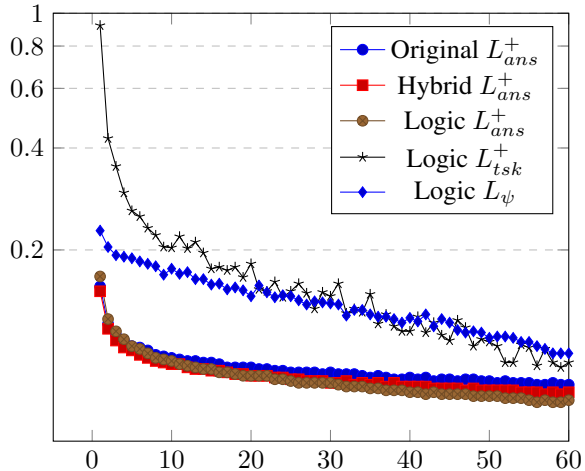
\begin{figure}[htbp]
    \centering
    \begin{tikzpicture}
        \begin{axis}[
            xmax=60,
            ymax=1.0,
            ymode=log,
            log ticks with fixed point,
            xtick={0,10,20,30,40,50,60},
            ytick={0,.2,.4,.6,.8,1.0},
            legend pos=north east,
            ymajorgrids=true,
            grid style=dashed,
        ]
        
        \addplot table[x=epoch,y=trainAnswerLoss-Original] {loss_plot.dat};
        \addlegendentry{Original $L^{+}_{ans}$}
        \addplot table[x=epoch,y=trainAnswerLoss-Hybrid] {loss_plot.dat};
        \addlegendentry{Hybrid $L^{+}_{ans}$}
        \addplot table[x=epoch,y=trainAnswerLoss-Logic] {loss_plot.dat};
        \addlegendentry{Logic $L^{+}_{ans}$}
        \addplot table[x=epoch,y=trainTaskLoss-Logic] {loss_plot.dat};
        \addlegendentry{Logic $L^{+}_{tsk}$}        
        \addplot table[x=epoch,y=trainLogicLoss-Logic] {loss_plot.dat};
        \addlegendentry{Logic $L_{\psi}$}                
        \end{axis}
    \end{tikzpicture}
    \caption{Plots (h-axis: epoch index, v-axis: loss value) of multi-task losses in log-scale for \textbf{Original}, \textbf{Hybrid}, \textbf{Logic} experiments}
    \label{fig:result-1}
\end{figure}

The proposed architecture for $f_{tsk}$ comprises a single 16-D fully connected (FC) layer between input question vectors $d$ and an output label of $\vert \mathcal{T} \vert = 48$ different semantic tasks. A soft-max operator normalizes an output of the 16-D FC layer for computing the cross-entropy supervised loss w.r.t. provided labels of semantic tasks in training phases. This minor addition in the overall architecture does not affect training, validation and testing procedures, originally proposed in MAC network \cite{hudson2019gqa} except a few extra gradient computation and back-propagation for supervisedly learning semantic task recognition and updating respective weights of the FC layer.

\subsection{Experimental Results}

\begin{table}[htbp]
\centering
\begin{tabular}{|l|l|l|l|}
\hline
Metrics      & \multicolumn{1}{p{1cm}|}{\textbf{Original}} & \multicolumn{1}{p{1cm}|}{\textbf{Hybrid}} & \multicolumn{1}{p{1cm}|}{\textbf{Logic}} \\ \hline
Accuracy     & 55.80\% & 55.47\% & \textbf{56.01\%} \\ \hline
Consistency  & 74.54\% & \textbf{77.49\%} & \textbf{79.21\%} \\ \hline
Distribution & 6.43    & 6.25             & \textbf{5.78} \\ \hline
\end{tabular}
\caption{Performance comparison between \textbf{Original}, \textbf{Hybrid} and \textbf{Logic} experiments on MAC network in Accuracy, Consistency and Distribution metrics. For the distribution metric, the lower the better. For other metrics, the higher the better.}
\label{tab:result-1}
\end{table}

To illustrate effect of the proposed hybrid batch and logic consistent loss, we carry out three experiments named \textbf{Original}, \textbf{Hybrid} and \textbf{Logic}. In the \textbf{Original} experiment, the models are trained with randomly selected batch; meanwhile, the \textbf{Hybrid} and \textbf{Logic} are trained with hybrid batch, previously described. The loss function of the 'Original' and 'Hybrid' experiments have only one term $L^{+}_{ans}$, a single-task learning of correct answer. Finally, the \textbf{Logic} experiment is a tri-task learning of correct answers, semantic tasks and consistency between the semantic tasks $L^{+}_{ans} + L^{+}_{tsk} + L_{\psi}$ of which the consistent logic formulae $\psi$ is Eqn. \ref{lgceqn:e_rule}. All testing / validation results of these experiments are carried out with batches of random questions from test-dev partition in the GQA dataset.

Table \ref{tab:result-1} show results across these three different experiments w.r.t. to different metrics recommended by Hudson et al. \cite{hudson2019gqa}. Though the full-set of recommended metrics include 7 statistical metrics for the overall dataset and 10 different accuracy for subsets according to categories of semantic tasks mentioned in Table \ref{tab:result-2} of the Supplementary Material, we focus on three most important metrics Accuracy, Consistency and Distribution, shown in Table \ref{tab:result-1} to argue for advantages of using hybrid-batch data arrangement and consistency-augmented losses. First is the Accuracy metric, the overall accuracy of predicted answers given ground-truth answers regardless of its semantic types or whether they are binary or open questions. Second is the 'Consistency' metric measuring accuracy performance over groups of entailed questions given provided entailment relations in the test-dev partition. The third metric is Distribution, a Chi-square distance between distributions of predicted and ground-truth answers. This 'Distribution' metric provides a measure of statistical consistency of predicted answers over the whole test-dev dataset. The smaller 'Distribution' metric is, the more similar between predicted and ground-truth distributions of answers are. Hence, the more consistency predicted answers as ground-truth answers are almost always consistent.
 
 Regarding to 'Accuracy' metric performance between three experiments in Table \ref{tab:result-1}, \textbf{Hybrid} is comparable to \textbf{Original}; meanwhile, \textbf{Logic}'s result is slightly better than both. This means introduction of hybrid batches, multi-task frameworks and the proposed consistency-augmented loss do not hinder the overall accuracy and even slightly improve the performance of VQA machine in this metric. In contrast to only slight improvement in 'Accuracy', we notice significant improvement nearly \textbf{3\%} in 'Consistency' metrics when hybrid-batches are used in training phases, \textbf{Hybrid} column, instead of batches of random samples, \textbf{Original} column of Table \ref{tab:result-1}. Further \textbf{2\%} improvement in 'Consistency' metric is spotted when the proposed loss function of consistency rule in Eqn \ref{lgceqn:e_rule} is added into the multi-task loss. In brief, roughly \textbf{5\%} improvement, shown in \textbf{Logic} experiment's Consistency metric, is spotted in comparison with \textbf{Original}'s experiment. Finally, the improvement in consistency of predicted outputs by this proposal is confirmed by 'Distribution' metric, statistical consistency measurement across the distributions of all 1845 unique answers. Table \ref{tab:result-1} shows that distribution of predicted answers produced in \textbf{Hybrid} is only slightly better than that of \textbf{Original}'s result; meanwhile, far more similarity between answers of \textbf{Logic}'s experiments and the ground-truth answers is spotted in comparison with those of \textbf{Original} or \textbf{Hybrid} experiments.
 

Multiple losses $L^{+}_{ans}$, $L^{+}_{a}$ and $L_{\psi}$ of \textbf{Original}, \textbf{Hybrid} and \textbf{Logic} experiments are plotted in Figure \ref{fig:result-1}. When comparing answer losses $L^{+}_{ans}$ of three experiments especially after initial 30 epochs, the answer loss of \textbf{Logic} converges at lower value than that value of \textbf{Hybrid}, which is in turn lower than the value of \textbf{Original}. Hence, MAC-network trained in \textbf{Logic} experiment fits the training data better than in \textbf{Hybrid} and \textbf{Original}. This observation is confirmed by Figures \ref{fig:result-2},\ref{fig:result-3} in the Supplementary Material of answer accuracy on the first data partition for training and validation as they show convergence of \textbf{Logic} Answer Accuracy toward higher value than these of \textbf{Hybrid} and \textbf{Original}.

\begin{table}[htbp]
\footnotesize
\centering
\begin{tabular}{|p{.9cm}|p{1.8cm}|p{.8cm}|p{.8cm}|p{.8cm}|p{.8cm}|}
\hline
Semantic Tasks & Questions & Answers & \textbf{Original} & \textbf{Hybrid} & \textbf{Logic} \\ \hline
queryRel     & What is under the animal the sandwich is to the left of?           & towel & paper   & napkin & paper \\ \hline
exist \newline RelTrue & Do you see any towels under the brown animal?                      & yes   & no      & yes    & yes   \\ \hline
exist \newline RelTrue & Is there a towel under the cat that looks cream colored and white? & yes   & no      & no     & yes   \\ \hline
queryRel     & What is under the cat?                                             & towel & blanket & rug    & paper \\ \hline
queryRel     & What is under the brown cat?                                       & towel & blanket & rug    & paper \\ \hline
\end{tabular}
\caption{Answers and Predictions of \textbf{Original}, \textbf{Hybrid} and \textbf{Logic} experiments for a question '05451380' and its entailed questions}
\label{tab:result-3}
\end{table}

Table \ref{tab:result-3} shows anecdotal performance and differences between \textbf{Original}, \textbf{Hybrid} and \textbf{Logic} experiments. In this example, \textbf{Logic} has the best accuracy as it answers both questions with 'existRelTrue' semantic tasks correctly and consistently; meanwhile, \textbf{Original} and \textbf{Hybrid} experiments give none and one correct answer. Though \textbf{Logic} gave wrong answers to all three questions about queryRel, it demonstrates its consistency by giving the same answer to three different questions with the same semantic task. Meanwhile, \textbf{Original}  and \textbf{Hybrid} change their answers despite that three questions are about the same object. \textbf{Logic} may predict wrong answers but it is at least consistent in comparison with \textbf{Original} and \textbf{Hybrid}.



\section{Conclusion}
We address the inconsistency in GQA by introducing semantic structures and consistency-augmented losses into training processes. Our experiments show that  the family structures help increase consistency in VQA performances. The consistency is further improved when logically consistent loss function is introduced in the framework of multi-task learning. Remarkably, this consistency-augmented loss is model-agnostic and applicable to any existing gradient-based VQA framework as long as structural relationships exist between questions in training data.

\bibliographystyle{plain}
\bibliography{references}
\clearpage
\appendix

\section{Additional Results}

\begin{table}[htbp]
\centering
\begin{tabular}{|l|l|l|l|}
\hline
Metrics      & \multicolumn{1}{p{1cm}|}{\textbf{Original}} & \multicolumn{1}{p{1cm}|}{\textbf{Hybrid}} & \multicolumn{1}{p{1cm}|}{\textbf{Logic}} \\ \hline
Binary       & 69.13\% & 69.36\% & \textbf{69.91\%} \\ \hline
Open         & 43.30\% & 42.45\% & 42.99\% \\ \hline
Accuracy     & 55.80\% & 55.47\% & \textbf{56.01\%} \\ \hline
Consistency  & 74.54\% & \textbf{77.49\%} & \textbf{79.21\%} \\ \hline
Validity     & 95.08\% & 95.07\%          & 94.99\% \\ \hline
Plausibility & 91.35\% & 91.23\%          & 91.12\% \\ \hline
Distribution & 6.43    & 6.25             & \textbf{5.78} \\ \hline
choose       & 69.18\% & 67.98\%          & 67.26\% \\ \hline
compare      & 60.26\% & 61.17\%          & 63.44\% \\ \hline
logical      & 73.18\% & 73.96\%          & 74.83\% \\ \hline
query        & 43.30\% & 42.45\%          & 42.99\% \\ \hline
verify       & 68.07\% & 68.73\%          & 69.58\% \\ \hline
attr         & 55.95\% & 56.34\%          & 57.70\% \\ \hline
cat          & 55.43\% & 54.21\%          & 54.68\% \\ \hline
global       & 63.78\% & 62.94\%          & 62.61\% \\ \hline
obj          & 76.20\% & 75.94\%          & 76.20\% \\ \hline
rel          & 50.08\% & 49.40\%          & 49.52\% \\ \hline
\end{tabular}
\caption{Performance comparison between \textbf{Original}, \textbf{Hybrid} and \textbf{Logic} experiments on MAC network in the extended set metrics}
\label{tab:result-2}
\end{table}

Metric 'Binary' measures accuracy of answer prediction for questions with yes/no answer as does metric 'Open' for questions with other answers. Meanwhile, metric 'Accuracy' measures answer accuracy for all questions in test-dev partition of GQA. Metric 'Consistency' measures accuracy over a group of entailed questions, pre-defined relations between question samples in test-dev. Metric 'Validity' checks whether a given answer is within a question scope e.g. answers of 'red', 'green' or other colors to a color question; meanwhile, metric 'Plausibility' checks sensibility of an answer given a question (e.g. cats usually do not swim in a lake) by examining occurrence of answers w.r.t. question topics across the whole data set. For instances, red and green are plausible colour for apples but purple as implausible. Metric 'distribution' measures a Chi-Square statistical difference between true and predicted distributions of answers. Hence, the smaller this metric is, the better a model manage to capture conditional answer distribution. Metrics 'choose', 'compare', etc represent answer accuracy for different categorical semantics of questions.

When comparing \textbf{Original} and \textbf{Hybrid} experiments in Table \ref{tab:result-1}, we can see small difference in almost all metrics except 'Consistency'. The consistency, accuracy across groups of entailed questions, is almost \textbf{3\%} higher when structured data are provided and duo-task loss $L^{+}_{ans} + L^{+}_{tsk}$ are employed. Involving the proposed logic loss with the mentioned consistent logic formulae further improves the consistency by \textbf{2\%}. Overall, there is \textbf{5\%} in consistency score between the proposed \textbf{Logic} and \textbf{Original} experiment. Furthermore, consistent logic loss produce predicted answers of which distribution is much closer to the ground-truth answer distribution and improve the overall accuracy, as shown in metrics 'Distribution' and 'Accuracy' of Table \ref{tab:result-1}.

\begin{figure}[htbp]
    \centering
    \begin{tikzpicture}
        \begin{axis}[
            xmax=60,
            ymax=1.0,
            log ticks with fixed point,
            xtick={0,10,20,30,40,50,60},
            ytick={0,.1,.2,.3,.4,.5,.6,.7,.8,.9,1.0},
            legend columns=2,
            legend style={at={(0.5,-0.1)},anchor=north},
            ymajorgrids=true,
            grid style=dashed,
        ]
        
        \addplot table[x=epoch,y=trainAnswerAcc-Original] {acc_plot.dat};
        \addlegendentry{Original Answer}
        \addplot table[x=epoch,y=trainAnswerAcc-Hybrid] {acc_plot.dat};
        \addlegendentry{Hybrid Answer}
        \addplot table[x=epoch,y=trainAnswerAcc-Logic] {acc_plot.dat};
        \addlegendentry{Logic Answer}
        \addplot table[x=epoch,y=trainTaskAcc-Logic] {acc_plot.dat};
        \addlegendentry{Logic Task}
        \end{axis}
    \end{tikzpicture}
    \caption{Plots (h-axis: epoch index, v-axis: accuracy) of train accuracy for \textbf{Original}, \textbf{Hybrid} and \textbf{Logic} experiments}
    \label{fig:result-2}
\end{figure}
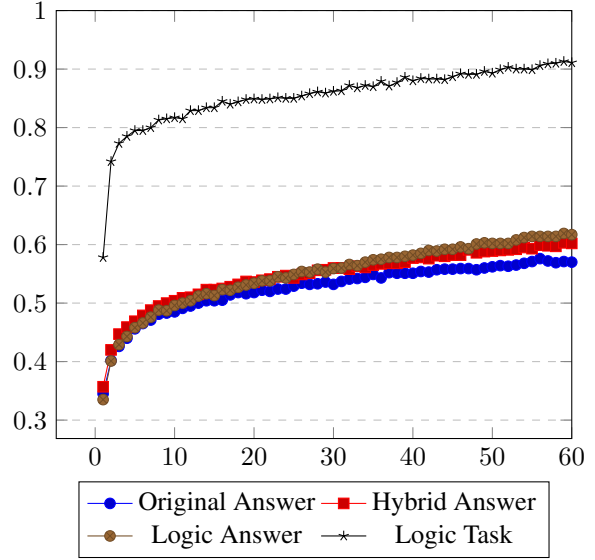

\begin{figure}[htbp]
    \centering
    \begin{tikzpicture}
        \begin{axis}[
            xmax=60,
            ymax=1.0,
            log ticks with fixed point,
            xtick={0,10,20,30,40,50,60},
            ytick={0,.1,.2,.3,.4,.5,.6,.7,.8,.9,1.0},
            legend pos=north east,
            ymajorgrids=true,
            grid style=dashed,
        ]
        \addplot table[x=epoch,y=valAnswerAcc-Original] {acc_plot.dat};
        \addlegendentry{Original Answer}
        \addplot table[x=epoch,y=valAnswerAcc-Hybrid] {acc_plot.dat};
        \addlegendentry{Hybrid Answer}
        \addplot table[x=epoch,y=valAnswerAcc-Logic] {acc_plot.dat};
        \addlegendentry{Logic Answer}        
        \end{axis}
    \end{tikzpicture}
    \caption{Plots (h-axis: epoch index, v-axis: accuracy) of validation accuracy on 1st data partition for \textbf{Original}, \textbf{Hybrid} and \textbf{Logic} experiments}
    \label{fig:result-3}
\end{figure}
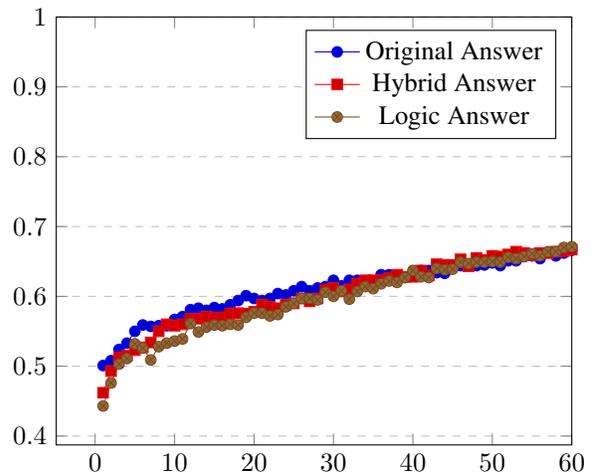

\section{Memory Attention and Compositional cell}
\label{apd:macell}

Hudson and Manning \cite{hudson2018compositional} pioneers a transparent and versatile design of neural networks for tackling Visual Question and Answer task. The MAC-network architecture has three main units: an input unit, a recurrent unit including a number of MAC cells and an output unit. The input unit uses \textit{d}-dimensional biLSTM image data, question into following embedded vectors: contextual words $cw_1,\ldots,cw_S$, a question or a query $q=[\overleftarrow{cw_1},\overrightarrow{cw_S}]$ and an image knowledge base $\mathbf{K}^{HxWxd}$. Subsequently, a vector $q$ is transformed into a position-aware vector $\mathbf{q_i}=W_{i}^{dx2d}q + b^{d}_{i}$
It is noted that $cw$ is an output state of biLSTM on $S$ words of the question and the arrows $[\leftarrow,\rightarrow]$ show whether the state is generated by a forward or backward direction of biLSTM. Furthermore, $H,W$ details how many vertical and horizontal regions are partitioned and encoded into d-dimensional embedded vectors. 

At the heart of MACnetwork is a series of MAC cells, a recurrent cell, reasoning about possible answers given the question representation $[\overleftarrow{cw_1},\overrightarrow{cw_S}]$, contextual words $cw_1,\ldots,cw_S$ and knowledge base $\mathbf{K}^{HxWxd}$ from the input unit. The number of MAC cells, $p$, represents the number of reasoning steps with dual hidden states $c_i$ and $m_i$ of dimension $d$ given the initial states $c_0$ and $m_0$. The control $c_i$ guides a cell's soft-attention to contextual words $cw_s; s=1,\ldots,S$. Meanwhile, the memory $m_i$ holds a immediate result of performing a control instruction $c_i$ on the knowledge-base vector $\mathbf{K}^{HxWxd}={\mathbf{k} \big|^{H,W}_{h,w=1,1} }$. This memory is an integration of the previous memory vector $m_{i-1}$ and new information $r_i$, extracted from a shift in attention toward a specific region $k_{h,w}$ facilitated by the current control signal $c_i$. 

Across multiple iterative reasoning steps, the control signal and memory states are read, manipulated and written by three operational units: a read-unit RU, a control-unit CU and a write-unit RU, inspired by design principles of computer organization. The control-unit (CU) is trained to identify a probable sequence of operation w.r.t. soft-attention to contextual word vectors ${cw_1,\ldots,cw_S}$; meanwhile, the read-unit (RU) utilises newly updated a control signal $c_i$ to guide its soft-attention to the respective knowledge-based vector and retrieve $r_i$. Finally, the write-unit (WU) integrates the retrieved information $r_i$ with the previous memory state $m_{i-1}$ to form an updated memory state $m_i$. Through operations of these three units, MAC cell regulates an interaction of information between textural modality of question space and visual modality of knowledge (image) space. The interaction is quantified through probability distribution only and the memory state is updated by indirect means such as soft-attention maps and sigmoidal gating mechanisms. Hudson et. al. \cite{hudson2019learning} emphasizes this stark difference between MAC-network approach from fusing question and image vectors together as they argued a strict separation between the representation spaces helps improve generalization and transparency of the solution. Further details about MAC-network as well as evidences for this claim can be found in \cite{hudson2019learning}.

\section{T-norm Fuzzy Logic}
\label{apd:tfl}
Binary logic with true value (1) and false value (0) is a rigid interpretation of logic and reasoning. Fuzzy logics or many-valued logics have been introduced to extend the absolute truth / false values ${0,1}$ to an admissible continuous close range of truth degree $[0,1]$, including the abosolutely true or false as boundary cases. The truth degree of fuzzy-logic variable corresponds to a real-unit interval $[0,1]$, usually an outcome of normalization after a softmax or sigmoid layer in Neural Network architecture. Hajek \cite{hajek2013metamathematics} defines a binary operation for this many-value truth degree between two fuzzy-logic variables upon a certain t-norm, a triangular norm, as follows.

\theoremstyle{definition}
\begin{definition}[t-norm]
An operation $T: [0,1]^2 \rightarrow [0,1]$ is a t-norm if and only if for every $x,y,z \in [0,1]$: \\
Commutativity: $T(x,y) = T(y,x)$ \\
Associativity: $T(T(x,y),z) = T(x,T(y,z))$ \\
Boundary: $T(x,1) = x, T(x,0) = 0$ \\
Non-decreasing: $ x < y \rightarrow T(x,z) < T(y,z)$ 
\end{definition}

The logical connectives are defined in terms of T-norm operators  $x \otimes y = T(x,y)$  as follows.

\begin{table}[htbp]
    \centering
    \begin{tabular}{c|c|p{4cm}}
         Name & Connective & T-norm  \\
         residual-impl & $ x \Rightarrow y$ & $max\{z: x \otimes z \leq y\}$ \\
         bi-residuum & $ x \Leftrightarrow y$ & $(x \Rightarrow y ) \otimes (y \Rightarrow x)$ \\
         weak-conj & $x \wedge y$ & $x \otimes ( x \Rightarrow y )$ \\
         weak-disj & $x \vee y$ & $ ((x \Rightarrow y) \Rightarrow y)$ \newline $\otimes ((y \Rightarrow x ) \Rightarrow x)$ \\
         residual-neg & $\sim x$ & $x \Rightarrow 0$ \\
         strong-neg & $\neg x$ & $1 - x$ \\
         t-conorm & $x \oplus y$ & $\neg(\neg x \otimes \neg y)$ \\
         material-imply & $x \rightarrow y$ & $\neg x \oplus y$
    \end{tabular}
    \caption{Logical connectives and its T-norm formulae}
    \label{tab:tnorm-1}
\end{table}

There are three fundamental continuous T-norms in literature as shown in Table \ref{tab:tnorm-2} in Supplementary Material, corresponding to different realization of T-norm operator i.e. Godel T-norm $T(x,y) =min\{x,y\}$, Lukasiewicz T-norm $T(x,y)=max\{0,x+y-1\}$ and Product T-norm $T(x,y)=x \cdot y$.

\begin{table}[htbp]
    \centering
    \begin{tabular}{c|p{2.1cm}|p{2.5cm}|p{4cm}}
         & Godel & Lukasiewicz & Product \\
        $x \otimes y$ & $\min\{x,y\}$ & $\max\{0,x+y-1\}$ & $x\cdot y$ \\
        $x \Rightarrow y$ & $x\leq y?1:y$ & $\min\{1,1-x+y\}$ & $ x \leq y ? 1 : \frac{y}{x}$ \\
        $x \Leftrightarrow y$ & $x\leq y ? x : y $ & $1-\vert x-y \vert $ & $x=y?1:$ \newline $\min\{\frac{x}{y},\frac{y}{x}\}$ \\
        $x \wedge y$ & $\min\{x,y\}$ & $\min\{x,y\}$ & $\min\{x,y\}$ \\
        $x \vee y $ & $\max\{x,y\}$ & $\max\{x,y\}$ & $\max\{x,y\}$ \\
        $ \sim x $ & $x=0?1:0$ & $1-x$ & $x=0?1:0$ \\
        $ \neg x $ & $1-x$ & $1-x$ & $1-x$ \\
        $ x \oplus y $ & $\max\{x,y\}$ & $\min\{1,x+y\}$ & $x+y-x \cdot y$ \\
        $ x \rightarrow y $ & $\max\{1-x,y\}$ & $\min\{1,1-x+y\}$ & $1-x+x \cdot y$
    \end{tabular}
    \caption{Three principal T-norm interpretation of logical connective}
    \label{tab:tnorm-2}
\end{table}

Table \ref{tab:tnorm-2} shows computation of truth degree of a logic operator i.e. residual imply $\rightarrow$ or residual negation $\sim x$ vary w.r.t. T-norm choide. However, these fundamental are specific of examples more generalized and parameterized \textit{Archimedean} t-norms, which can be constructed by unary monotone functions - generators \cite{klement2013triangular}. Klement et. al. \cite{klement2004triangular} describes this parameterized construction of T-norm in the following theorem.

\theoremstyle{plain}
\begin{theorem}{(generator)}
Let $g: [0,1] \rightarrow [0,+\infty]$ be a strictly decreasing function with $g(1)=0$ and $g(x)+g(y) \in [0,+\infty] \cup [g(0^{+}),+\infty]$ for all $x,y$ in $[0,1]$ and $g^{-1}$ its pseudo-inverse. The the function $T: [0,1]^{2} \rightarrow [0,1]$ defined as
\[
T(x,y) = g^{-1}\big( \min \{ g(0^{+}), g(x) + g(y)\}\big)
\]
is a t-norm and g is an additive generator for T. T is strict if $g(0^{+})=+\infty$ otherwise T is nilpotent. 
\label{def:tnorm-1}
\end{theorem}

Lets assume a generator function $g(x)=1-x$, then it defines the Lukasiewicz T-norm $T_L$ \[T_L(x,y)=1-\min\{1,1-x+1-y\}=\max\{0,x+y-1\}\]
If the generator is defined as $g(x)=-\log(x)$, then we get Product T-norm $T_P$ as \[T_P(x,y) = \exp\Big(-\big(\min\{+\infty,-\log(x)-\log(y)\}\big)\Big)=x\cdot y\]
The definition \ref{def:tnorm-1} of T-norm with additive generator $g$ leads to subsequent interpretations of other logical connections such as.
\begin{align}
    &x \Rightarrow y = g^{-1}\Big(\max\{0,g(y)-g(x)\}\Big) \\
    &y \Leftrightarrow y = g^{-1}\Big(\vert g(x)-g(y) \vert\Big) \\
    &x \oplus y = 1-g^{-1}\Big(\min \big\{ g(0^{+}),g(1-x)+g(1-y) \big\} \Big)
\end{align}

The generator function $g$ can be parameterised with a parameter $\lambda$ to generate different T-norm i.e. $T_G,T_L,T_P$; two prominent classes of parameterised generators are defined below.

\theoremstyle{definition}
\begin{definition}{\textbf{The Schweize-Sklar family.}} 
For $\lambda \in (-\infty,+\infty)$, consider:
\[
g^{SS}_{\lambda}(x) = \begin{cases} -\log(x) & \text{if } \lambda = 0 \\ \frac{1-x^{\lambda}}{\lambda} & \text{otherwise} \end{cases}
\]
The corresponding Scheweizer-Sklar t-norms are defined accordingly to:
\[
T^{SS}_{\lambda}(x,y) = 
\begin{cases}
T_G(x,y) & \text{if } \lambda = -\infty \\
(x^{\lambda}+y^{\lambda}-1^{\frac{1}{\lambda}} & \text{if } -\infty < \lambda < 0 \\
T_P(x,y) & \text{if } \lambda = 0 \\
\max\{0,x^{\lambda}+y^{\lambda}-1\}^{\frac{1}{\lambda}} & \text{if } 0 < \lambda < +\infty \\
T_{D}(x,y) & \text{if} \lambda = +\infty 
\end{cases}
\]
This t-norm is strictly decreasing for $\lambda \geq 0$ and continuous with respect to $\lambda \in [-\infty,+\infty]$. In addition, $T^{SS}_1 = T_{L}$.
\end{definition}.
\theoremstyle{definition}
\begin{definition}{\textbf{Frank family.}} For $\lambda \in [0,+\infty]$, consider:
\[
g^{F}(x)=
\begin{cases}
-\log(x) & \text{if} \lambda = 1 \\
1-x & \text{if} \lambda = +\infty \\
\log\Big(\frac{\lambda-1}{\lambda^{x}-1}\Big) & \text{if otherwise} 
\end{cases}
\]
The corresponding Frank t-norms are defined below.
\[
T^{F}_{\lambda} = 
\begin{cases}
T_G &\text{if } \lambda = 0 \\
T_P &\text{if } \lambda = 1 \\
T_L &\text{if } \lambda = +\infty \\
\log_{\lambda}\Big( 1 + \frac{(\lambda^x-1)(\lambda^y-1)}{\lambda-1}\Big) &\text{otherwise}
\end{cases}
\]
The overall class of Frank t-norms is decreasing and continuous w.r.t. $\lambda \in [-\infty,+\infty]$.
\end{definition}


\begin{table*}[htbp]
\centering
\begin{tabular}{|l|l|l|}
\hline
"allDiffFalse"      & "existAttrOrTrue" & "twoDiffTrue"        \\ \hline
"allDiffTrue"       & "existAttrTrue"   & "twoSameFalse"       \\ \hline
"allSameFalse"      & "existFalse"      & "twoSameTrue"        \\ \hline
"allSameTrue"       & "existNotFalse"   & "verifyAttrAndTrue"  \\ \hline
"chooseAttr"        & "existNotTrue"    & "verifyAttrFalse"    \\ \hline
"chooseGlobal"      & "existOrFalse"    & "verifyAttrTrue"     \\ \hline
"chooseObj"         & "existOrTrue"     & "verifyAttrsFalse"   \\ \hline
"chooseRel"         & "existRelFalse"   & "verifyAttrsTrue"    \\ \hline
"common"            & "existRelTrue"    & "verifyGlobalFalse"  \\ \hline
"compare"           & "existTrue"       & "verifyGlobalTrue"   \\ \hline
"existAndFalse"     & "queryAttr"       & "verifyRelFalse"     \\ \hline
"existAndTrue"      & "queryAttrObj"    & "verifyRelTrue"      \\ \hline
"existAttrFalse"    & "queryGlobal"     & "queryNotObj"        \\ \hline
"existAttrNotFalse" & "queryObj"        & "existNotOrTrue"     \\ \hline
"existAttrNotTrue"  & "queryRel"        & "existNotOrFalse"    \\ \hline
"existAttrOrFalse"  & "twoDiffFalse"    & "verifyAttrAndFalse" \\ \hline
\end{tabular}
\caption{A list of 48 semantic tasks, defined by entailment graphs and used in this work}
\label{tab:semtsk}
\end{table*}

\begin{table*}[htbp]
\centering
\begin{tabular}{|l|l|}
\hline
"$\forall x_1, \forall x_2$ queryObj($x_1$) $\Rightarrow$ queryAttrObj($x_2$)"                 & "$\forall x_1, \forall x_2$ existAttrNotFalse($x_1$) $\Rightarrow$   existAttrFalse($x_2$)"        \\ \hline
"$\forall x_1, \forall x_2$ queryAttrObj($x_1$) $\Rightarrow$ existAttrTrue($x_2$)"            & $\forall x_1, \forall x_2$ existNotOrFalse($x_1$) $\Rightarrow$   existNotFalse($x_2$)"           \\ \hline
"$\forall x_1, \forall x_2$ existAttrTrue($x_1$) $\Rightarrow$   existAttrOrTrue($x_2$)"         & $\forall x_1, \forall x_2$ existNotOrFalse($x_1$) $\Rightarrow$   existAttrNotFalse($x_2$)"       \\ \hline
"$\forall x_1, \forall x_2$ existAttrTrue($x_1$) $\Rightarrow$   existNotTrue($x_2$)"            & "$\forall x_1, \forall x_2$ existNotOrFalse($x_1$) $\Rightarrow$   existAttrOrFalse($x_2$)"        \\ \hline
"$\forall x_1, \forall x_2$ existAttrTrue($x_1$) $\Rightarrow$   existAttrNotTrue($x_2$)"        & "$\forall x_1, \forall x_2$ existAttrOrFalse($x_1$) $\Rightarrow$   existAttrFalse($x_2$)"         \\ \hline
"$\forall x_1, \forall x_2$ existAttrOrTrue($x_1$) $\Rightarrow$   existNotOrTrue($x_2$)"        & "$\forall x_1, \forall x_2$ verifyAttrsTrue($x_1$) $\Rightarrow$   verifyAttrTrue($x_2$)"          \\ \hline
"$\forall x_1, \forall x_2$ existNotOrTrue($x_1$) $\Rightarrow$   existOrTrue($x_2$)"            & "$\forall x_1, \forall x_2$ verifyAttrAndTrue($x_1$) $\Rightarrow$   verifyAttrTrue($x_2$)"         \\ \hline
"$\forall x_1, \forall x_2$ existOrTrue($x_1$) $\Rightarrow$   existTrue($x_2$)"                 & "$\forall x_1, \forall x_2$ verifyAttrTrue($x_1$) $\Rightarrow$   queryAttr($x_2$)"                \\ \hline
"$\forall x_1, \forall x_2$ queryAttrObj($x_1$) $\Rightarrow$   queryObj($x_2$)"                 & "$\forall x_1, \forall x_2$ queryAttr($x_1$) $\Rightarrow$   verifyAttrFalse($x_2$)"               \\ \hline
"$\forall x_1, \forall x_2$ queryNotObj($x_1$) $\Rightarrow$   existNotTrue($x_2$)"              & "$\forall x_1, \forall x_2$ queryAttr($x_1$) $\Rightarrow$   chooseAttr($x_2$)"                    \\ \hline
"$\forall x_1, \forall x_2$ existNotTrue($x_1$) $\Rightarrow$   existTrue($x_2$)"                & "$\forall x_1, \forall x_2$ verifyAttrFalse($x_1$) $\Rightarrow$   verifyAttrAndFalse($x_2$)"      \\ \hline
"$\forall x_1, \forall x_2$ existAttrNotTrue($x_1$) $\Rightarrow$   existTrue($x_2$)"            & "$\forall x_1, \forall x_2$ verifyAttrAndFalse($x_1$) $\Rightarrow$   chooseAttr($x_2$)"           \\ \hline
"$\forall x_1, \forall x_2$ existAndTrue($x_1$) $\Rightarrow$   existTrue($x_2$)"                & "$\forall x_1, \forall x_2$ chooseAttr($x_1$) $\Leftrightarrow$   chooseObj($x_2$)"         \\ \hline
"$\forall x_1, \forall x_2$ existRelTrue($x_1$) $\Rightarrow$   existTrue($x_2$)"                & "$\forall x_1, \forall x_2$ verifyGlobalTrue($x_1$) $\Rightarrow$   verifyGlobalFalse($x_2$)"      \\ \hline
"$\forall x_1, \forall x_2$ existRelTrue($x_1$) $\Leftrightarrow$   verifyRelTrue($x_2$)" & "$\forall x_1, \forall x_2$ verifyGlobalTrue($x_1$)   $\Leftrightarrow$ queryGlobal($x_2$)" \\ \hline
"$\forall x_1, \forall x_2$ verifyRelTrue($x_1$) $\Leftrightarrow$   queryRel($x_2$)"     & "$\forall x_1, \forall x_2$ verifyGlobalFalse($x_1$) $\Rightarrow$   chooseGlobal($x_2$)"          \\ \hline
"$\forall x_1, \forall x_2$ verifyRelTrue($x_1$) $\Leftrightarrow$   chooseRel($x_2$)"    & "$\forall x_1, \forall x_2$ compare($x_1$) $\Rightarrow$   common($x_2$)"                          \\ \hline
"$\forall x_1, \forall x_2$ existOrFalse($x_1$) $\Rightarrow$   existFalse($x_2$)"               & "$\forall x_1, \forall x_2$ common($x_1$) $\Rightarrow$   twoSameTrue($x_2$)"                      \\ \hline
"$\forall x_1, \forall x_2$ existFalse($x_1$) $\Rightarrow$   existNotFalse($x_2$)"              & "$\forall x_1, \forall x_2$ twoSameTrue($x_1$) $\Leftrightarrow$   twoDiffFalse($x_2$)"     \\ \hline
"$\forall x_1, \forall x_2$ existFalse($x_1$) $\Rightarrow$   existAttrNotFalse($x_2$)"          & "$\forall x_1, \forall x_2$ twoSameFalse($x_1$) $\Leftrightarrow$   twoDiffTrue($x_2$)"     \\ \hline
"$\forall x_1, \forall x_2$ existFalse($x_1$) $\Rightarrow$   existRelFalse($x_2$)"              & "$\forall x_1, \forall x_2$ allSameTrue($x_1$) $\Leftrightarrow$   allDiffFalse($x_2$)"     \\ \hline
"$\forall x_1, \forall x_2$ existFalse($x_1$) $\Rightarrow$   existAndFalse($x_2$)"              & "$\forall x_1, \forall x_2$ allSameFalse($x_1$) $\Leftrightarrow$   allDiffTrue($x_2$)"     \\ \hline
"$\forall x_1, \forall x_2$ existNotFalse($x_1$) $\Rightarrow$   existAttrFalse($x_2$)"          &                                                                 \\ \hline
\end{tabular}
\caption{A list of 51 FOL logic formulae of semantic links, defined by entailment graphs and used in this work}
\label{tab:semlog}
\end{table*}

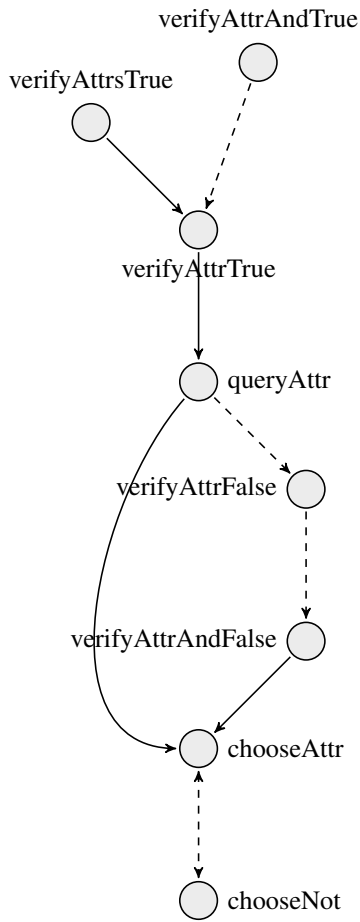
\begin{figure}[t]
    \centering
    \begin{tikzpicture}[modal]
        \node[world, label={above:verifyAttrsTrue}] (verifyAttrsTrue) {};
        \node[world, label={above:verifyAttrAndTrue}, yshift=.8cm, xshift=.2cm] (verifyAttrAndTrue) [right=of verifyAttrsTrue] {};
        \node[world, label={below:verifyAttrTrue}] (verifyAttrTrue) [below right=of verifyAttrsTrue] {};
        \node[world, label={right:queryAttr}] (queryAttr) [below =of verifyAttrTrue] {};
        \node[world, label={left:verifyAttrFalse}] (verifyAttrFalse) [below right=of queryAttr] {};
        \node[world, label={left:verifyAttrAndFalse}] (verifyAttrAndFalse) [below=of verifyAttrFalse] {};
        \node[world, label={right:chooseAttr}] (chooseAttr) [below left=of verifyAttrAndFalse] {};
        \node[world, label={right:chooseNot}] (chooseNot) [below=of chooseAttr] {};
        \path[->] (verifyAttrsTrue) edge[] (verifyAttrTrue);
        \path[->] (verifyAttrAndTrue) edge[dashed] (verifyAttrTrue);
        \path[->] (verifyAttrTrue) edge[] (queryAttr);
        \path[->] (queryAttr) edge[dashed] (verifyAttrFalse);
        \path[->] (verifyAttrFalse) edge[dashed] (verifyAttrAndFalse);
        \path[->] (verifyAttrAndFalse) edge[] (chooseAttr);
        \path[<->] (chooseAttr) edge[dashed] (chooseNot);
        \path[->] (queryAttr) edge[bend right=40,in=-90] (chooseAttr);
    \end{tikzpicture}
    \captionof{figure}{Entailment graph about attributes of subjects / objects}
    \label{fig:family-batch-2}
\end{figure}

\begin{table}[t]
    \centering
    \begin{tabular}[b]{|p{1.3cm}|p{4.2cm}|l|}
        \hline
        Semantic Task & Questions & Answers \\ \hline
        Verify \newline AttrsTrue  &  Do these bananas look peeled and ripe?  &  yes \\ \hline
        Verify \newline AttrTrue & Do the bananas that are not unpeeled look ripe?  &  yes \\ \hline
        Verify \newline AttrFalse & Are these unripe bananas?  &  no \\ \hline
        Verify \newline AttrsFalse & Are the ripe bananas unpeeled and cooked? & no \\ \hline   
        Choose \newline Attr  &  Do the bananas that look ripe look peeled or unpeeled?  &  peeled \\ \hline
    \end{tabular}  
    \captionof{table}{Examples of questions and answers for Figure \ref{fig:family-batch-2}}
    \label{tab:family-batch-2}      
\end{table}

\end{document}